# Unsupervised Spatio-Temporal State Estimation for Fine-grained Adaptive Anomaly Diagnosis of Industrial Cyber-physical Systems

Haili Sun[a,1], Yan Huang[b,1], Lansheng Han[a,*], Cai Fu[a], Chunjie Zhou[b]

[a] Hubei Key Laboratory of Distributed System Security, Hubei Engineering Research Center on Big Data Security, School of Cyber Science and Engineering, Huazhong University of Science and Technology, Wuhan, 430074, China

[b] National Key Laboratory of Science and Technology on Multispectral Information Processing, School of Artificial Intelligence and Automation, Huazhong University of Science and Technology, Wuhan, 430074, China

## Abstract

Accurate detection and diagnosis of abnormal behaviors such as network attacks from multivariate time series (MTS) are crucial for ensuring the stable and effective operation of industrial cyber-physical systems (CPS). However, existing researches pay little attention to the logical dependencies among system working states, and have difficulties in explaining the evolution mechanisms of abnormal signals. To reveal the spatio-temporal association relationships and evolution mechanisms of the working states of industrial CPS, this paper proposes a fine-grained adaptive anomaly diagnosis method (i.e. MAD-Transformer) to identify and diagnose anomalies in MTS. MAD-Transformer first constructs a temporal state matrix to characterize and estimate the change patterns of the system states in the temporal dimension. Then, to better locate the anomalies, a spatial state matrix is also constructed to capture the inter-sensor state correlation relationships within the system. Subsequently, based on these two types of state matrices, a three-branch structure of series-temporal-spatial attention module is designed to simultaneously capture the series, temporal, and space dependencies among MTS. Afterwards, three associated alignment loss functions and a reconstruction loss are constructed to jointly optimize the model. Finally, anomalies are determined and diagnosed by comparing the residual matrices with the original matrices. We conducted comparative experiments on five publicly datasets spanning three application domains (service monitoring, spatial and earth exploration, and water treatment), along with a petroleum refining simulation dataset collected by ourselves. The results demonstrate that MAD-Transformer can adaptively detect fine-grained anomalies with short duration, and outperforms the state-of-the-art baselines in terms of noise robustness and localization performance.

---

1 Equal contribution;

* Corresponding Author: Lansheng Han; E-mail address: hanlansheng@hust.edu.cn (Lansheng Han), hailisun@hust.edu.cn (Haili Sun);

## 1. Introduction

With the rapid development of technology, complex cyber-physical systems (CPSs) have become ubiquitous in contemporary industrial fields. These systems integrate computation and physical processes, realize the seamless integration and flow of information, thereby greatly improving production efficiency, system performance, and resource utilization. In various industries, such as smart manufacturing, intelligent transportation, and smart energy management, CPSs are gradually replacing traditional systems and becoming an important driving force for industrial upgrading and transformation. The widespread application of complex CPSs brings unprecedented opportunities for changes in the current industrial fields. CPSs achieve intelligent control of the production process by real-time monitoring of various equipment (such as sensors) states, product quality, and environmental parameters on the production line. Based on data analysis and artificial intelligence algorithms, these systems can predict equipment failures, optimize production processes, reduce resource waste, and significantly enhance production efficiency and product quality.

However, these CPSs are highly vulnerable to malicious cyber-attacks, with severe consequences ensuing. For example, on March 7, 2019, Venezuela's power grid suffered a cyber-attack, resulting in a nationwide blackout that caused more than $800 million in economic losses and 46 deaths [1]. These CPSs continuously generate large amounts of multivariate time series (MTS) data, such as the readings from sensors monitoring flow rates, temperatures, and liquid levels in water treatment plants. Therefore, researchers have proposed methods based on MTS analysis to detect attack behaviors in CPSs.

Sensor errors, data transmission fluctuations, system variability, and external environmental interference frequently introduce noise into the MTS data generated by CPS. This necessitates algorithms with strong noise resistance capabilities to ensure high accuracy and low false alarm rates (FAR) in anomaly detection, which are crucial for guaranteeing the normal operation of CPSs and mitigating substantial high economic losses. Moreover, to facilitate operators to quickly and accurately identify the causes and locations of anomalies, it is usually necessary to precisely locate the anomalies, that is, to determine which sensors cause the corresponding anomalies.

In real-world scenarios, the fault-tolerance mechanism of modern systems implies that short-term anomalies caused by time fluctuations or system state transitions may not ultimately lead to real system failures. Therefore, if an assessment of the severity of the anomalies can be provided, it will help the operators find a suitable trade-off between the impact of the anomalies and the losses caused by downtime. After all, industrial systems are different from traditional IT systems, and their downtime can also cause serious economic losses. It is reported that a one-minute downtime of a car manufacturer may result in up to $20,000 in economic losses [2]. This paper assumes that the severity of the anomalies is proportional to the number of anomalous devices (such as sensors and actuators) and the duration of the anomalies, as shown in Figure 1, there are two anomalies A1 and A2. It can be seen that A1 has 5 anomalous points and A2 has 3 anomalous points, and A1 lasts longer than A2. Therefore, we consider A1 to be more severe than A2.

In constructing anomaly detection and diagnosis algorithms for industrial fields, one of the first challenges we encounter is the scarcity of labeled samples, and in some cases, there may not be any labeled samples at all. In such situations, due to the insufficient number of training samples and interference from noise, traditional supervised learning algorithms may lead to overfitting or underfitting issues, thereby greatly reducing the model's generalization capability. In recent years, some researchers have proposed anomaly detection algorithms based on unsupervised learning, such as statistical-based methods [3], distance-based methods [4], clustering-based methods [5][6], density estimation methods [7][8], et al. Although they can make initial judgments on the anomalies in CPS systems, most of them overlook the temporal association present in MTS data, failing to capture the temporal evolution patterns of the time series themselves.

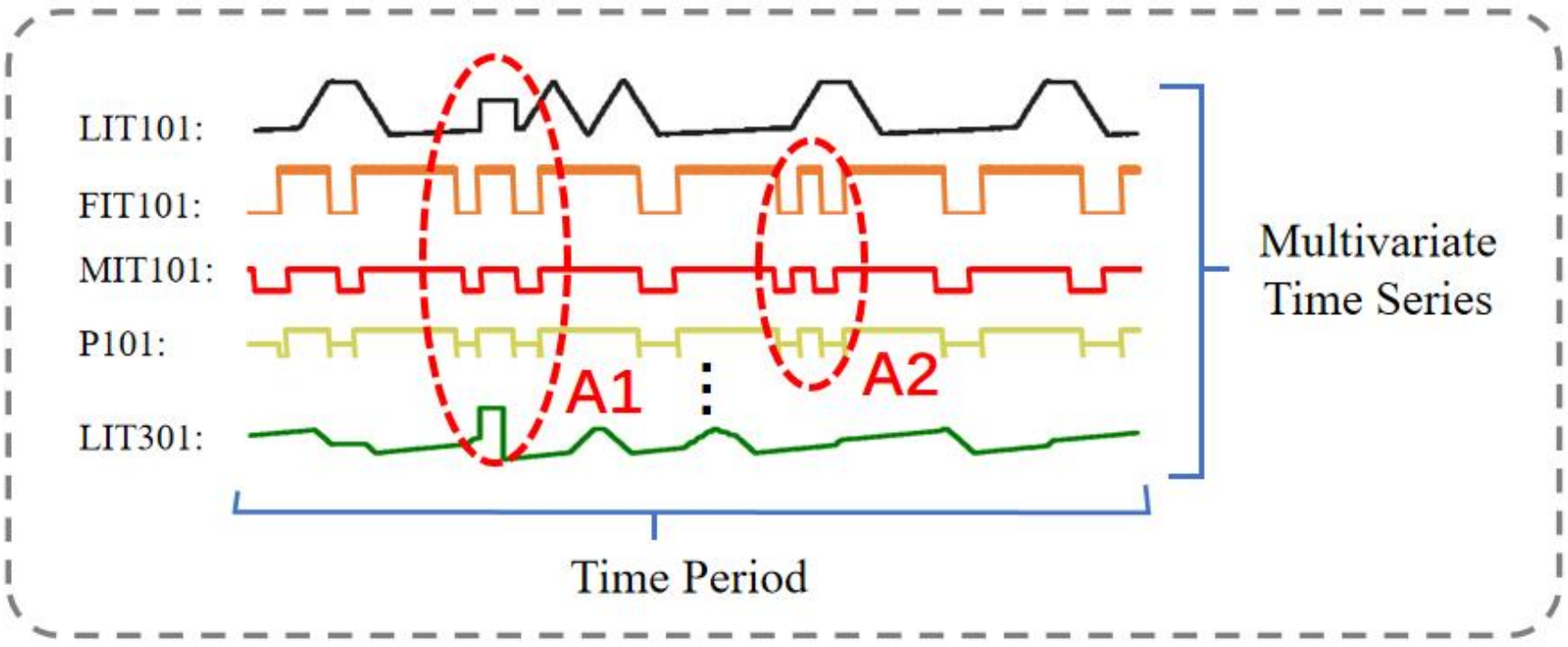

Figure 1. Explanation of anomaly severity: A1 and A2 are two anomalies in the SWAT dataset. A1 is more severe than A2, as it causes more anomalous points and lasts longer. Therefore, we consider A1 to have greater damage and higher severity than A2.

Additionally, although some deep learning models based on LSTM are capable of capturing temporal dependencies within time series, they overlook the spatial dependencies that exist within the business domain of the time series. In the real world, providing operators with the location of anomalies and corresponding exception result assessments based on the severity of events is meaningful. However, existing anomaly detection methods for MTS rarely provide assessments of the severity of anomalies. Although MSCRED [9] can capture spatial dependencies within MTS to determine the most likely locations of anomalies, it ignores the temporal dependencies between them, and its diagnosis of anomaly severity is relatively coarse-grained.

Accurate detection and diagnosis of anomalous behaviors such as network attacks from multivariate time series is crucial for ensuring the stable and effective operation of industrial information-physical systems. To this end, not only do we require meticulously designed models that capture the temporal correlations within MTS, but also a learning mechanism that recognizes the spatial business correlations between different time series data. Additionally, it is necessary to localize anomalous results and adaptively assess their severity at a fine-grained level, to enhance the interpretability of anomalies and accelerate the response speed of operators. Existing research has paid less attention to the logical dependencies within the system's working state, resulting in an incomplete explanation of the evolution mechanism of system anomaly signals.

To capture both the temporal and spatial business dependencies among multivariate time series and to locate and diagnose the severity of anomalies with fine granularity, this paper proposes a novel unsupervised fine-grained anomaly detection and diagnosis method to ensure the safe operation of industrial systems. Specifically, we use a temporal state matrix to depict the overall system state at different time steps, capturing the temporal dependencies in MTS. Meanwhile, we use another spatial state matrix to capture the spatial dependencies among time series. Then, these two state matrices and the original time series are fed into a specially designed

MAD-Transformer with three-branch attention, to explicitly learn the temporal, spatial, and sequential dependencies in multivariate time series. Next, we design three correlation difference functions to align these three dependencies in the feature space. Finally, based on the reconstructed sequence, temporal, and spatial matrices, we compute the anomaly scores and the corresponding residual matrices. The anomaly scores are used to determine whether a time point is anomalous or not, the spatial residual matrix is used to identify the most likely location of the anomalies, and the temporal residual matrix is used to diagnose the severity of the anomalies with fine granularity.

In general, our main contributions are:

(1) We propose an unsupervised anomalous pattern recognition model, MAD-Transformer, for detecting and diagnosing anomalies in CPS systems. This model characterizes the evolution patterns of anomalous signals in CPS systems from both temporal and spatial dimensions. Through fine-grained health state estimation and anomaly localization, it ensures the stable and secure operation of industrial systems.

(2) The temporal and spatial business state dependencies between MTS are modeled separately as temporal state matrices and spatial state matrices. Additionally, a three-branch structure attention module, MAD-Attention, is designed to extract correlative association patterns from the original time series, temporal evolution state matrices, and business spatial state matrices. Three alignment losses are formulated to align these three types of association patterns, thereby extracting more informative features.

(3) By computing the residuals between the reconstructed time and space feature matrices and the original matrix, it is possible to identify the most likely location of anomalies and to assess the severity of anomalies with fine-grained precision. Furthermore, the severity of anomalies can also be evaluated adaptively at a granular level based on the data's business context.

(4) Extensive comparative experiments were conducted on five public datasets from three applications (service monitoring, space, and earth exploration, and water treatment) and one self-built petroleum refining simulation dataset. The results demonstrated the superior performance of the proposed method compared to 21 benchmark models, as well as its accuracy in anomaly localization and its capability for adaptive fine-grained anomaly diagnosis.

## 2. Related Work

Unsupervised MTS anomaly detection is an important and challenging real-world problem. Recently, researchers have proposed various methods for this task. Based on the criterion for anomaly determination, these methods can be roughly classified into distance-based, density estimation-based, clustering-based, prediction-based, and reconstruction-based methods.

Distance-based methods are a traditional type. For example, the k-Nearest Neighbors (KNN) algorithm [4] calculates the anomaly score for each data sample based on the average distance to its k nearest neighbors. For density estimation-based methods, classical methods are local outlier factor (LOF) [10] and connectivity outlier factor (COF) [11], which identify outliers by calculating local density and local connectivity, respectively. DAGMM [8] and MPPCACD [12] integrate Gaussian mixture models to estimate the density of representations. Clustering-based methods compute anomaly scores by calculating the distance from samples to cluster centers is used as the anomaly score. SVDD [13] and Deep SVDD [14] gather representations of normal data into a compact cluster. THOC [15] fuses multi-scale temporal features from intermediate layers using a hierarchical clustering mechanism and detects anomalies based on multi-layer distances. ITAD [16] conducts clustering on decomposed tensors.

Despite the effectiveness of the above three types of methods in various applications, they may not handle multivariate time series well, as they cannot capture temporal dependencies among time series. To address this issue, many deep learning-based methods have started employing ARIMA or LSTM to model the temporal dependencies. These methods can be roughly categorized into prediction-based and reconstruction-based methods.

The method based on prediction predicts the value of the next time step by predicting the residual and discriminates anomalies based on the prediction error. VAR [17] extends ARIMA and predicts the future based on lagged correlation covariance. Hundman et al. [18] demonstrated the feasibility of long short-term memory (LSTM) in detecting spacecraft anomalies and introduced a method for dynamically setting thresholds without relying on annotations. Similarly, Tariq et al. [19] proposed a data-driven satellite anomaly detection algorithm based on LSTM, which simultaneously uses neural networks and probabilistic clustering to identify anomalies from telemetry data.

Reconstruction-based methods detect anomalies by calculating reconstruction losses between reconstructed samples and real samples. Park [20] proposed an LSTM-VAE model, which employs an LSTM backbone for temporal dependencies modeling and utilizes a Variational Autoencoder (VAE) for reconstruction. Su et al. [21] further extended the LSTM-VAE and constructed OmniAnomaly, which adopted a normalized flow and utilized reconstruction probabilities for anomaly detection. Li et al. [22] developed InterFusion which utilized a hierarchical VAE as a backbone. They modeled both the inter-dependencies and intra-dependencies among multiple sequences. Based on adversarial training of autoencoders, Audibert et al. [23] proposed USAD, which aims to identify anomalies from multivariate time series in a fast and stable manner. Additionally, recently popular Generative Adversarial Networks (GAN) [24] are also used for constructing reconstruction-based anomaly detection models [25][26][27]. Although these two methods achieve better generalization capabilities than traditional approaches, they solely capture temporal dependencies at the time step level, thus limiting their ability to detect rare anomalies in multivariate time series.

Except for the aforementioned methods, other works [28][29][30] dynamically construct inter-sequence graphs across time and perform graph prediction or reconstruction tasks. For example, EvoNet [28] extracts representative MTS segments as nodes and learns their transition probabilities during the training process. However, it aims to capture the temporal pattern changes between different periods, rather than the inter-sequence relationship changes. MSCRED [9] uses three signature matrices of different scales to model the inter-correlations among multivariate time series and evaluates anomaly severity by calculating the residuals of these matrices. However, its diagnosis results are not only slightly coarse but also increase the computational and storage overhead. Chen et al. [31] use dynamic graphs to model the dependencies among multivariate time series, but they cannot diagnose and locate anomalies in essence.

Unlike existing models that only model temporal or sequential dependencies separately, we propose state matrices to represent the temporal and spatial dependencies in a multivariate time series segment and construct a three-branch attention structure MAD-Attention, to capture the sequential, temporal, and spatial correlations among time series. As a result, we achieve a complementary characterization extraction of the multivariate time series from three dimensions:

sequence, time, and space, which greatly enhances the information of extracted features. Moreover, we also designed three alignment correlation functions to align these three correlations in feature space to obtain more expressive representations. Finally, by calculating residuals between the reconstructed matrices and input matrices, we can identify anomalies, locate their probable positions, and diagnose their severity at a fine-grained level.

# 3. Motivation and Definition

## 3.1 Motivation Description

To capture the underlying states for effective and interpretable anomaly identification, an intuitive approach is to treat a sequence of time series as possible patterns and then model their sequential dependencies [26][27]. However, industrial time series are often accompanied by noise, and there are complex business-related inter-series relationships, which are useful for characterizing the system's states [30], i.e. spatial business association information.

To this end, graph-based methods [29][30] can be employed to model the structural dependencies between time series. However, these methods often fail to provide a severity diagnosis of anomalies, which is necessary to assist operators in taking appropriate responses. Therefore, in this work, we propose a novel temporal and spatial association state matrix to describe the relationships between multivariate time series, and we design a three-branch attention module to capture the temporal evolution patterns of these state matrices. Finally, we reconstruct these state matrices and calculate the reconstruction error between the input matrix and the reconstructed matrix to differentiate, locate, and diagnose anomalies.

Intuitively, in the real world, there may be multiple anomalies and the duration of the anomalies may be longer than one time step, and the model may not be able to reconstruct the state matrices that deviate from the normal working mode of the system well.

Specifically, we propose a temporal state matrix to depict the relationship between the system's working state and its temporal evolution. We believe that when a system is operating normally, the changes in its state over time should also conform to certain working patterns. This transforms the modeling of the time series itself into modeling the changes in the system's state over time, which can be more effective in distinguishing anomalies and identifying the duration of anomalies. Additionally, in the real world, locating the position where anomalies occur is of significant importance for rapid response. Therefore, to locate the position of anomalies, we have also designed a spatial state matrix to depict the dependencies between sensors, thereby identifying the most likely location of anomaly occurrence.

It is noteworthy that in the spatial state matrix, each row represents a sensor, so the more rows that are anomalous, the more locations the anomaly has occurred. Furthermore, previous works [32][33] have shown that the interrelationships between the system's sensors help characterize the local states of the system. Therefore, this paper proposes to depict the system's state evolution over time from both time and space perspectives, enabling fine-grained diagnosis and localization of anomalies.

## 3.2 Problem Statement

Given a segment of the MTS $x = [x_1, x_2, ..., x_T] \in R^{n \times T}$, which consists of $n$ features collected in $T$ time steps, as the training set. Let $x_t = [x_t^1, x_t^2, ..., x_t^n] \in R^{n \times 1}$ denotes the $n$ features at time step $t$, and $x_t^i$ represents the value of the $i$-th feature at time step $t$; $x^i =$

$[x_1^i, x_2^i, ..., x_w^i] \in R^{1\times w}$ denote the values of the $i$-th feature in a time window with length $w$. The test set is also an MTS with $n$ features, which is collected within a time range different from that of the training set. The goal of this paper is to detect abnormal behaviors from MTS, and to locate and diagnose the severity of the abnormal results, providing one or more most likely positions of the abnormality, as well as the ability to assess the severity (duration) of the abnormality at any granularity of time scale.

## 4. Methodology

### 4.1 Overall Architecture

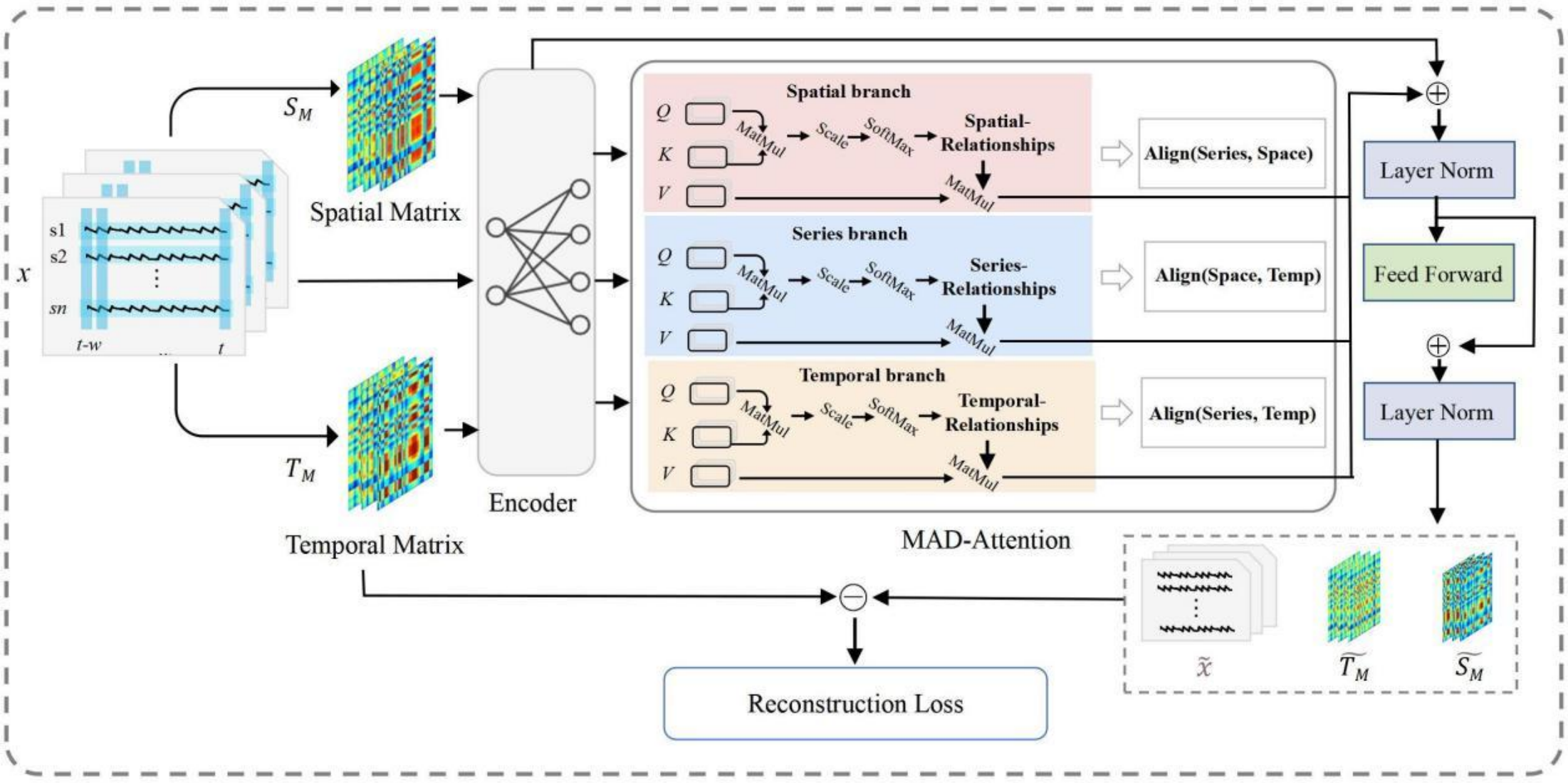

Figure 2. The overall architecture of the proposed fine-grained anomaly detection model MAD-Transformer. Firstly, spatial state matrices $S_M$ and temporal state matrices $T_M$ are constructed for MTS $x$. Three-layer convolution neural network encoders are used to encode features for, $x$ and $T_M$ respectively. Then, a three-branch attention mechanism (i.e. MAD-Attention) is employed to separately learn the associations among sequences, time, and space, and to generate reconstructed time series signals $\tilde{x}$ and reconstructed matrices $\widetilde{S_M}$ and $\widetilde{\mathrm{T_M}}$. The model trains the network using alignment loss and reconstruction loss based on the series-temporal-spatial associations, predicts the anomalous locations, and assesses the severity of anomalies through the residual values of the reconstructed state matrices.

Figure 2 illustrates our proposal for a multi-branch Transformer to uncover more meaningful associations. By learning the temporal and spatial state changes, we model the normal operational patterns of the system, enabling more precise detection of anomalous behaviors. Technically, we introduce a three-branch attention mechanism (MAD-Attention) to separately learn the series-temporal-spatial associations. Along with this, we design three alignment losses to align these three associations in the feature space, thereby obtaining more expressive features.

To capture the temporal and spatial dependencies among MTS, this paper first constructs a temporal state matrix $T_M$ by calculating the correlations between pairwise time steps such as ($x_{t1}$ and $x_{t2}$) along the time dimension. Then, to extract the spatial dependencies, along the feature dimension, we also build a feature correlation matrix $S_M$ by calculating the correlations between pairwise features such as ($x^i$ and $x^j$). Subsequently, the $T_M$, $S_M$ and the original time series are fed into a three-branch attention structure, MAD-Attention, to train a model that models the normal pattern of the MTS from three dimensions: time, feature, and space. Finally, the trained model is used to detect abnormal behaviors from the test set and provide one or more possible locations of the anomalies, as well as assess the severity (duration) of the anomalies in any granularity of time scale.

## 4.2 Modeling Spatio-Temporal Dependencies

Unlike existing methods that only model dependencies between time series, the multi-branch Transformer model proposed in this paper captures two types of association relationship information from the raw sequences and the proposed state matrices: spatial association and temporal association.

The historical working status of the system will affect the current observed values of the system. For instance, in the process of water treatment, if the previous state was drainage, then the current readings of the water tank's pressure sensor will decrease. Conversely, if the state was filling, the readings would be opposite. However, the filling state must precede the drainage state.

Intuitively, in industrial control systems, a change in the value of one device (sensor) can cause a change in the values of other related devices. In other words, the relationships among sensors (devices) are crucial. For instance, in a water treatment plant, if the level sensor readings increase, the flow rate sensor readings may decrease (the validity of this example needs to be confirmed).

The spatial influence mentioned above can naturally be represented by a spatial state matrix, which is primarily composed of the associations between pairs of sensors. Meanwhile, the temporal association can be represented by a temporal state matrix, which determines how the system's working state evolves across different rows of the temporal state matrix.

**a) Temporal State Matrix**

To characterize the overall state change pattern of the system within a given period (e.g., an MTS segment of length $w$), we construct a $w * w$ temporal state matrix $T_M$ based on the inner products of the time series pairs corresponding to each pair of time steps within this time segment. Specifically, given an MTS of length $w$, $x = [x_{t-w}, x_{t-w+1}, ..., x_t]$, where $x_i = [x_i^1, x_i^2, ..., x_i^n]$ denotes the MTS corresponding to time step $i$, the calculation formula for the state relationship $T_{ij} \in T_M$ between time steps $i$ and $j$ is as follows:

$$T_{ij} = \frac{x_i \circ x_j}{\tau_t} \tag{1}$$

where $\circ$ denotes the dot product, $\tau_t$ is the hyperparameter of the temporal state matrix. $n$ is the dimension of the time series (i.e., the number of sensors in the system), and $T_M$ refers to the temporal dynamic association matrix.

As a result, each row of this matrix represents the correlation relationship between the corresponding time point sequence and all other time points. However, when an anomaly occurs in the system, it disrupts this correlation relationship, represented by a matrix image and time series waveform. The more rows in the matrix that exhibit anomalies, the longer the duration of the anomaly. Therefore, by modeling this temporal state matrix, it is not only possible to detect anomalies but also to characterize their duration at any time granularity, which is beneficial for later fine-grained diagnosis of the severity of anomalies.

**b) Spatial State Matrix**

In the real world, a system often comprises hundreds or even thousands of devices, hence simply providing alarm information for anomalies is insufficient. It is also necessary to indicate the possible location of the anomaly, enabling operations personnel to quickly locate and investigate the cause, thereby reducing the losses associated with extended system downtime. Intuitively, the relationships between devices (sensors) within the system should conform to certain patterns when the system is operating normally. Therefore, to locate anomalies, we have characterized the association state patterns between sequences in MTS and constructed a spatial state matrix $S_M$ with dimension $n * n$. Specifically, for a given MTS $x = [x_{t-w}, x_{t-w+1}, \ldots, x_t]$ of length $w$, the association relationship $S_{ij} \in S_M$ between any two time series $x_w^i = (x_{t-w}^i, x_{t-w+1}^i, \ldots, x_t^i)$ and $x_w^j = (x_{t-w}^j, x_{t-w+1}^j, \ldots, x_t^j)$ can be defined as their dot product:

$$S_{ij} = \frac{x_w^i \circ x_w^j}{\tau_s} \tag{2}$$

where $\circ$ denotes the dot product, $\tau_s$ is the hyperparameter of the spatial state matrix. This spatial state matrix characterizes the association relationships between any two different time series within the system, thereby enabling the capture of potential anomalies. This is because once certain time series exhibit anomalies, they are bound to violate the association relationships that hold under normal conditions.

## 4.3 MAD-Transformer

Due to the inability of traditional Transformers to perform time series anomaly detection, we propose the MAD-Transformer, designed to detect and diagnose anomalies in MTS, as shown in Figure 2 (model structure diagram). The MAD-Transformer is characterized by alternating stacks of multi-branch attention blocks and feed-forward layers. This stack structure facilitates the learning of latent correlations from deep, multi-level features.

Given a segment of MTS $x \in R^{w \times n}$ (where $w$ is the length of the time series and $n$ is the dimension of the time series), we first calculate its corresponding temporal state matrix $T_M$ and spatial state matrix $S_M$ separately according to Section 4.2. Assuming the model has $K$ layers, the formulas for the $k$-th layer can be formalized as follows:

$$H^k = Layer_Norm(MAD_Attention([x, T_M, S_M]^{k-1}) + [x, T_M, S_M]^{k-1}) \tag{3}$$

$$[x, T_M, S_M]^k = Layer_Norm(MLP(H^k) + H^k) \tag{4}$$

where $x^k$, ${T_M}^k \in R^{w * d_{channel}}$, ${S_M}^k \in R^{n * d_{channel}}$, $k \in \{1,2, \ldots, K\}$ denotes the output of the $k$-th layer with $d_{channel}$ channels. The input $[x, T_M, S_M]^0 = Encoder([x, T_M, S_M])$ is initialized to represent the embedded original sequence. $H^k$ denotes the intermediate hidden state at layer $k$. $MAD_Attention(\cdot)$ is used to calculate the associative relationships between different embedding.

**MAD-Attention:** Since the single-branch self-attention mechanism [34] cannot simultaneously model sequence association, temporal association, and spatial association, we propose a time-space attention with a three-branch structure. Within this structure, series association is used to learn sequence-level dependencies from the original input, allowing for the adaptive discovery of the most effective sequence associations. For temporal association, we first compute a temporal state matrix based on the relationships between pairs of time points, and then extract the most effective temporal associations from this matrix. Similarly, for spatial association, we calculate a corresponding spatial state matrix from the original sequence based on the relationships between pairs of sensors (dot product), which represents the original spatial associations, and then adaptively find the most distinguishing spatial associations from this matrix. Please note that these three types of associations simultaneously maintain the sequence association, temporal association, and spatial association of the time series, providing more information than point-by-point representations.

As shown in Figure 2, MAD-Attention consists of three branches: the Series branch, the Temporal branch, and the Space branch. The Series branch is responsible for learning the sequence associations within the time series. The Temporal branch captures temporal association relations from the temporal state matrix. Similarly, the Space branch explicitly learns the spatial correlations between sensors from the spatial state matrix. These three branches are complementary, reflecting the temporal and inter-sensor associations within the time series. We believe that the combination of these three shall result in more discriminative representations, as violating any of these dependencies would lead to anomalies. The output of the $k$-th layer of MAD-Attention is given by Equation 5:

$$\begin{aligned} Initiallization: Q_x, K_x, V_x &= x^{k-1}w_{Q_x}^k, x^{k-1}w_{K_x}^k, x^{k-1}w_{V_x}^k \\ Q_{T_M}, K_{T_M}, V_{T_M} &= {T_M}^{k-1}w_{QTM}^k, {T_M}^{k-1}w_{K_{TM}}^k, {T_M}^{k-1}w_{V_{TM}}^k \\ Q_{S_M}, K_{S_M}, V_{S_M} &= {S_M}^{k-1}w_{QSM}^k, {S_M}^{k-1}w_{K_{SM}}^k, {S_M}^{k-1}w_{V_{SM}}^k \\ Series\ Association: S^k &= \mathrm{Softmax}\left(\frac{Q_x(K_x)^T}{\sqrt{d_{model}}}\right) \\ Temporal\ Association:\ Temp^k &= \mathrm{Softmax}\left(\frac{Q_{T_M}(K_{T_M})^T}{\sqrt{d_{model}}}\right) \\ Spatial\ Association: Space^k &= \mathrm{Softmax}\left(\frac{Q_{S_M}(K_{S_M})^T}{\sqrt{d_{model}}}\right) \\ Reconstruction: \widetilde{H}_x^k &= S^k V_x \\ \widetilde{H}_{TM}^k &= Temp^k V_{T_M} \\ \widetilde{H}_{SM}^k &= Space^k V_{S_M} \end{aligned} \tag{5}$$

where $Q_x, K_x, V_x, Q_{T_M}, K_{T_M}, V_{T_M} \in R^{w*d_{channel}}, Q_{S_M}, K_{S_M}, V_{S_M} \in R^{n*d_{channel}}$ denotes query, key, and value of the self-supervised sequence, time branch, and space branch, respectively. $w_{Q_x}^k, w_{K_x}^k, w_{V_x}^k, w_{QTM}^k, w_{K_{TM}}^k, w_{V_{TM}}^k, w_{QSM}^k, w_{K_{SM}}^k, w_{V_{SM}}^k \in R^{d_{channel}*d_{channel}}$ denotes parameter matrices of $\mathrm{Q_x, K_x, V_x, Q_{TM}, K_{TM}, V_{TM}, Q_{SM}, K_{SM}, V_{SM}}$ at the k -th layer respectively. $S^k, Temp^k \in R^{w*w}, Space^k \in R^{n*n}$ denotes sequence association, temporal association and spatial association respectively. The Softmax function normalizes the attention map along the last dimension. Therefore, in the aforementioned attention map ($S^k, Temp^k, Space^k$), each row conforms to a discrete distribution. $\widetilde{H}_x^k, \widetilde{H}_{TM}^k \in R^{w*d_{channel}}, \widetilde{H}_{SM}^k \in R^{n*d_{channel}}$ denote the intermediate representation of the $k$ -th layer after MAD-Attention. We denote Equation 5 as $MAD_attention(\cdot)$.

## 4.4 Loss Measurement

### 4.4.1 Association Alignment Loss

We utilize the symmetric KL divergence to align pairwise associations, which also represents the information gain between these two distributions. We average the association differences across multiple layers, combining the associations of multi-layer features into a more informative measurement as follows:

$$Align(Seri, Temp) = [\frac{1}{K}\sum_k^K (KL(Seri_{i,:}^k \| Temp_{i,:}^k) + KL(Temp_{i,:}^k \| Seri_{i,:}^k))]_{i=1,\dots,w} \quad (6)$$

$$Align(Seri, Space) = [\frac{1}{K}\sum_k^K (KL(Seri^k || Space^k) + KL(Space^k || Seri^k))] \quad (7)$$

$$Align(Temp, Space) = [\frac{1}{K}\sum_k^K (KL(Space^k || Temp^k) + KL(Temp^k || Space^k))] \quad (8)$$

$$Align_{total}(Seri, Temp, Space) = ||AssAlign(Seri, Temp)||_1 + ||AssAlign(Seri, Space)||_1 + ||AssAlign(Temp, Space)||_1 \quad (9)$$

In Equation 6, $KL(\cdot\|\cdot)$ denotes the KL divergence computed between two discrete distributions corresponding to each row for $i$ and $j$. Therefore, $AssAlign(Seri, Temp) \in R^{w*1}$ denotes the point-wise association differences between the multi-layer sequence association $S$ and the temporal association. The $i$-th element of the result corresponds to the $i$-th time point of the input sequence $X$. In Equations 7 and 8, $KL(\cdot\|\cdot)$ denotes the association differences between two distributions, as the dimensions of spatial association and sequence association as well as temporal association are different. $||\cdot||_1$ denotes the L1 norm.

#### 4.4.2 Reconstruction Loss

As an unsupervised task, we employ reconstruction loss to optimize the model. We calculate the reconstruction loss for the original sequence $x$, the temporal matrix $T_M$, and the spatial matrix $S_M$ separately. These reconstruction losses will guide the various associations to find the most informative message.

$$L_R = ||x - \tilde{x}||_F^2 + ||T_M - \widetilde{T_M}||_F^2 + ||S_M - \widetilde{S_M}||_F^2 \quad (10)$$

where $\tilde{x}$, $\widetilde{T_M}$ and $\widetilde{S_M}$ denotes the reconstructions of sequence x, temporal state matrix $T_M$ and spatial state matrix $S_M$. $||.||_F$ denotes Frobenius norm.

Additionally, note that since the temporal matrix and the spatial (feature) matrix respectively reflect the association relationships of the time series in the temporal dimension and the spatial (feature) dimension, once there is an abnormal sequence at a certain moment, it will inevitably violate at least one of these associations. Therefore, the purpose of reconstructing both here is to visualize the duration of an ongoing anomaly, as well as to identify which features (sensors) exhibit abnormalities, in order to determine the location of the anomaly and visualize its severity.

#### 4.4.3 Total Loss

The overall loss function consists of two parts: reconstruction loss $L_R$ and association alignment loss $Align_{total}(Seri, Temp, Space)$. The reconstruction loss guides the three associations of sequence, temporal and spatial to find the most informative association. Meanwhile, to further amplify the differences between anomalous and normal time points, we also employ an additional difference loss to align these three associations. This allows the three associations to influence and complement each other, enabling the learning of a more comprehensive pattern of normal data. As a result, the reconstruction of anomalies becomes more challenging, making anomalies more easily recognizable. According to Equation 9 and Equation 10, the overall loss function for the input sequence $x \in R^{w \times n}$ can be formulated as:

$$L_{Total} = L_R + \lambda * Align_{total}(S, T, F) \quad (11)$$

where $\tilde{x} \in R^{N \times d}$ is the reconstruction of $x$, $\|\cdot\|_F$ denotes Frobenius norm, λ is the hyperparameter used to balance the loss term.

## 4.5 Model Learning Strategy

We present the pseudo-code of MAD-Attention in ALGORITHM 1.

---

**ALGORITHM 1: MAD-Attention Mechanism (multi-head version)**

---

**Require:** input: $x \in R^{w \times d_{channel}}$; temporal matrix: $T_M \in R^{w \times d_{channel}}$; space matrix: $S_M \in R^{n \times d_{channel}}$

**Layer parameters:** linear projector for input and matrices: $x$Embedding, $Tm$Embedding, $Sm$Embedding linear projector for output: $x\text{Linear}_{output}$ $Tm\text{Linear}_{\text{output}}$, $Sm\text{Linear}_{output}$

$Q_x, K_x, V_x = Split(xEmbedding(x), dim = 1)$ $\quad \triangleright\ Q_x, K_x, V_x \in R^{w \times d_{channel}}$

$Q_{T_M}, K_{T_M}, V_{T_M} = Split(TmEmbedding(T_M), dim = 1)$ $\quad \triangleright\ Q_{T_M}, K_{T_M}, V_{T_M} \in R^{w \times d_{channel}}$

$Q_{S_M}, K_{S_M}, V_{S_M} = Split(SmEmbedding(S_M), \text{dim} = 1)$ $\quad \triangleright\ Q_{S_M}, K_{S_M}, V_{S_M} \in R^{n \times d_{channel}}$

**for** $(Q_x^l, K_x^l, V_x^l)$ **in** $(Q_x, K_x, V_x)$: $\quad \triangleright\ Q_x^l, K_x^l, V_x^l \in R^{w \times \frac{d_{channel}}{h}}$

$\quad S^l = softmax(\sqrt{\frac{h}{d_{channel}}} Q_x^l (K_x^l)^T)$ $\quad \triangleright\ S^l \in R^{w \times w}$

$\quad \widetilde{H}_x^l = S^l V_x^l$ $\quad \triangleright\ \widetilde{H}_x^l \in R^{w \times \frac{d_{channel}}{h}}$

$\widetilde{H}_x = x\text{Linear}_{output}(Concat([\widetilde{H}_x^1, \ldots, \widetilde{H}_x^h], dim = 1))$ $\quad \triangleright\ \widetilde{H}_x \in R^{w \times d_{channel}}$

**for** $(Q_{T_M}^l, K_{T_M}^l, V_{T_M}^l)$ **in** $(Q_{T_M}, K_{T_M}, V_{T_M})$: $\quad \triangleright\ Q_{T_M}^l, K_{T_M}^l, V_{T_M}^l \in R^{w \times \frac{d_{channel}}{h}}$

$\quad Temp^l = softmax(\sqrt{\frac{h}{d_{channel}}} Q_{T_M}^l (K_{T_M}^l)^T)$ $\quad \triangleright\ Temp^l \in R^{w \times w}$

$\quad \widetilde{H}_{T_M}^l = Temp^{l^l} V_{T_M}^l$ $\quad \triangleright\ \widetilde{H}_{T_M}^l \in R^{w \times \frac{d_{channel}}{h}}$

$\widetilde{H}_{T_M} = Tm\text{Linear}_{output}(Concat([\widetilde{H}_{T_M}^1, \ldots, \widetilde{H}_{T_M}^h], dim = 1))$ $\quad \triangleright\ \widetilde{H}_{T_M} \in R^{w \times d_{channel}}$

**for** $(Q_{S_M}^l, K_{S_M}^l, V_{S_M}^l)$ **in** $(Q_{S_M}, K_{S_M}, V_{S_M})$: $\quad \triangleright\ Q_{S_M}^l, K_{S_M}^l, V_{S_M}^l \in R^{n \times \frac{d_{channel}}{h}}$

$\quad Space^l = softmax(\sqrt{\frac{h}{d_{channel}}} Q_{S_M}^l (K_{S_M}^l)^T)$ $\quad \triangleright\ Space^l \in R^{n \times n}$

$\quad \widetilde{H}_{S_M}^l = Temp^{l^l} V_{S_M}^l$ $\quad \triangleright\ \widetilde{H}_{S_M}^l \in R^{n \times \frac{d_{channel}}{h}}$

$\widetilde{H}_{S_M} = Sm\text{Linear}_{output}\left(Concat([\widetilde{H}_{S_M}^1, \ldots, \widetilde{H}_{S_M}^h], dim = 1)\right)$ $\quad \triangleright\ \widetilde{H}_{S_M} \in R^{n \times d_{channel}}$

**return** $\widetilde{H}_x$, $\widetilde{H}_{T_M}$, $\widetilde{H}_{S_M}$

---

# 5. Anomaly Diagnosis

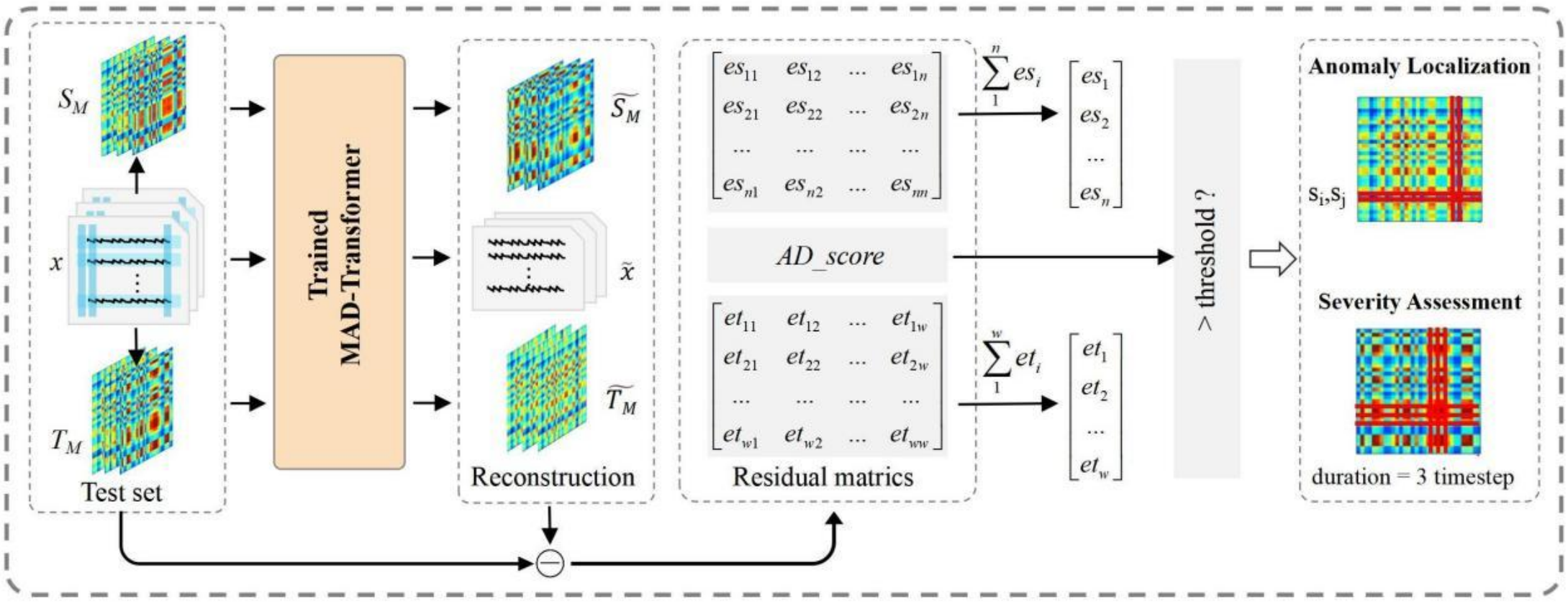

Figure 3: Anomaly Diagnosis Process. $S_M$ and $T_M$ denote the spatial state matrix and the temporal state matrix. $\tilde{x}$, $\widetilde{S_M}$, $\widetilde{T_M}$ denote the reconstructed time series, the reconstructed spatial state matrix and the reconstructed temporal state matrix. $et_{ij}$ denotes the error value at the $i$-th row and $j$-th column of the temporal state residual matrix. Similarly, $es_{ij}$ denotes the error value at the $i$-th row and $j$-th column of the spatial state residual matrix. $et_i$ and $es_i$ denote the errors at the $i$-th row of the temporal state residual matrix and the spatial state residual matrix, respectively. Subsequently, each row's error is compared with the corresponding threshold, and rows with errors greater than the threshold are identified as abnormal time steps or abnormal devices.

The process of anomaly detection and diagnosis is shown in Figure 3. Firstly, the original time series, time matrix, and spatial matrix are input into a trained model to obtain their reconstructions. Subsequently, the corresponding residual matrices are calculated based on the reconstructions and the original matrices. The sum of each row in the residual matrix yields the error determinant. Finally, this error determinant is compared with a threshold value, and any values greater than the threshold are considered anomalies.

## 5.1 Anomaly Detection

To accurately detect anomalies within the MTS data of CPS systems, we simultaneously leverage the merits of time series representation and distinguishable association differences. We measure the discrepancy between sequence correlation and temporal correlation at each time point, and calculate the anomaly score based on the normalized difference between sequence and temporal correlations. Therefore, for a given time series $x \in R^{w \times n}$, the final anomaly score is computed as follows:

$$AD_{score} = [||x_{t,:} - \tilde{x}_{t,:}||_2^2]_{t=1,\dots,w} \otimes Softmax(-Alig(Seri, Temp)) \quad (12)$$

where $\otimes$ represents element-wise multiplication. $AD_{score}$ denotes the point-by-point anomaly score in sample $x$. In the face of a better reconstruction, the anomaly may reduce the associated differences, yet it still yields a higher anomaly score. Therefore, this design allows the reconstruction error and the associated differences to work in collaboration, enhancing the accuracy of detection.

## 5.2 Anomaly Localization

The most likely locations of anomalies are determined by calculating the residual matrix of the spatial state matrix. Given the temporal state matrix $S_M$ and the reconstruction matrix $\widetilde{S_M}$, the residual matrix between them is calculated as:

$$Res_{S_M} = ||S_M - \widetilde{S_M}||_2^2 \otimes Softmax(-Alig(Seri, Space)) \quad (13)$$

The residuals for each row of $Res_{S_M}$ are summed as the final error for that row. Rows with errors greater than the threshold are considered to potentially correspond to devices that may have experienced anomalies.

## 5.3 Severity Assessment

As previously discussed, the severity of an anomaly is diagnosed by assessing the duration of the anomaly; the longer the duration, the more severe the anomaly is considered to be. Similar to the localization of anomalies, this diagnosis is carried out by computing the residual matrix between the temporal state matrix $T_M$ and its reconstruction matrix $\widetilde{T_M}$. Given the temporal state matrix $T_M$, the corresponding residual matrix is:

$$Res_{T_M} = ||T_M - \widetilde{T_M}||_2^2 \otimes Softmax(-Align(Seri, Temp)) \quad (14)$$

Subsequently, the residual matrix $Res_{T_M}$ is summed row-wise to serve as the final error for each row. Rows with errors exceeding the threshold are deemed to correspond to time points of anomalies. The greater the number of anomalous time points, the more severe the anomaly is considered to be.

It is worth noting that in practical applications, the granularity for assessing the severity of anomalies can be adaptively chosen based on the system's characteristics. For instance, each row of the temporal state matrix can represent the interval of one time step or a 10 second interval. Additionally, although our method provides three functions: anomaly identification, localization, and diagnosis, in real-world applications, it is possible to adaptively select one or two of these functions based on business requirements. However, we believe that combining the localization of anomalies with the diagnosis of the duration of anomalies to assess their severity results in a more comprehensive and meaningful judgment. Intuitively, the longer the duration of the anomaly and the more devices affected, the more severe it is likely to be. The process of anomaly diagnosis is illustrated in Figure 3.

# 6. Experiments

In this section, we conducted extensive comparative experiments with 21 benchmark models across 6 categories, including anomaly detection, anomaly localization, and severity assessment, based on 5 publicly available datasets for service monitoring, space and earth exploration, and water treatment applications, as well as a self-collected simulation dataset for water treatment. The goal was to verify whether the MAD-Transformer outperforms benchmark models in multi-variable time series anomaly detection tasks, whether it can localize the most likely location of anomalies, and whether it can provide effective fine-grained judgments of anomaly severity. Finally, we validated the impact of each component of the proposed model on its detection performance.

## 6.1 Datasets

**SWAT (Secure Water Treatment [36]):** Continuously collected over 14 days from a water treatment platform system containing 51 sensors and actuators.

**SMD (Server Machine Dataset [21]):** SMD is a dataset collected over five weeks from a large internet company, encompassing 38 dimensions.

**SMAP (Soil Moisture Active Passive satellite) and MSL (Mars Science Laboratory rover) [18]:** They consist of 55 and 25 dimensions, respectively, and include telemetry anomaly data from the spacecraft monitoring system including incidental spacecraft anomalies (ISA) reports.

**PSM (Pooled Server Metrics [37]):** Collected internally from multiple application server nodes at eBay, with a total of 26 dimensions.

**CTCS (Coupling Tank Control System)**: A self-collected dataset, derived from data gathered on our self-built petroleum refining simulation platform, encompasses seven dimensions. The physical diagram of the simulation platform is shown in Figure 4. It is worth noting that we extracted 20% from the training set to serve as the validation set.

Detailed statistics of the above datasets are shown in Table 1.

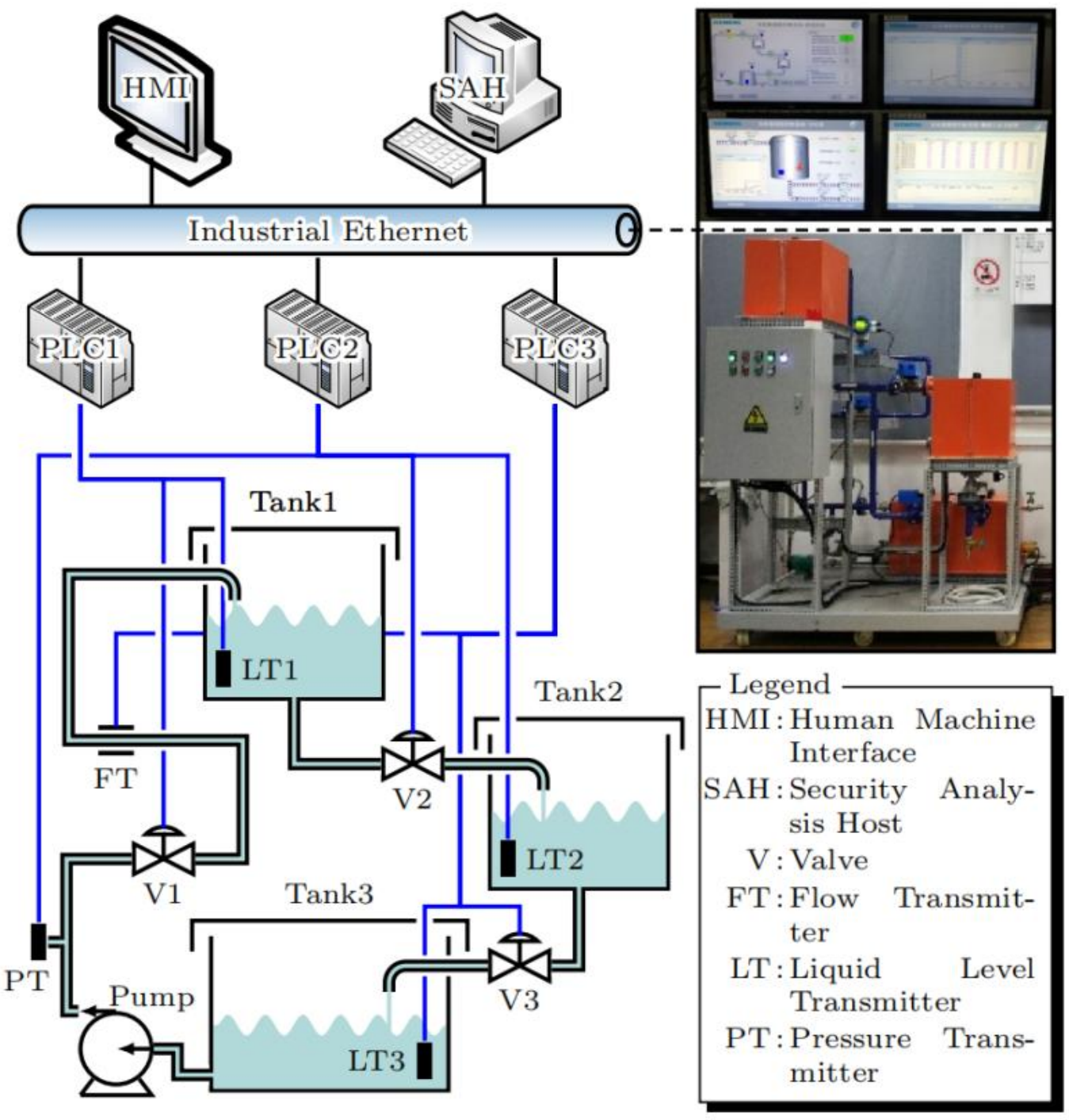

Figure. 4. Diagram of coupled tank control system in petroleum refining simulation platform

Table 1: Statistics for each dataset. Anomaly ratio refers to the proportion of abnormal data in the whole dataset.

| Dataset | SWAT | SMD | SMAP | MSL | PSM | CTCS |
|---|---|---|---|---|---|---|
| Applications | Water | Server | Space | Space | Server | Petroleum |
| Dimension | 51 | 38 | 25 | 55 | 25 | 25 |
| Window | 100 | 100 | 100 | 100 | 100 | 100 |
| Train | 495,000 | 708,405 | 135,183 | 58,317 | 132,481 | 30,000 |
| Test | 449,919 | 708,420 | 427,617 | 73,729 | 87,841 | 5,000 |
| Anomaly ratio (%) | 12.1 | 4.2 | 12.8 | 10.5 | 27.8 | 10.1 |

## 6.2 Baseline Methods

We compare the proposed model MAD-Transformer with 21 benchmark models across 6 categories: traditional classification models such as Random Forest [38] and One-Class SVM (OC-SVM) [7]; models based on reconstruction including ATransformer [35], MADGAN [26], USAD [23], InterFusion [22], BeatGAN [27], OMNIANOMALY [21], MSCRED [9],

LSTM-VAE [20] and TimesNet [39]; density estimation models such as DAGMM [8], MPPCACD [12] and LOF [10]; clustering-based models including ITAD [16], THOC [21] and Deep SVDD [14]; autoregressive-based models such as LSTM [18], CL-MPPCA [19] and VAR [17]; and graph-based models including MTAD-GAT [29] and GDN [30]. Among these, ATransformer [35] and GDN [30] are currently the best-performing models.

## 6.3 Evaluation Metrics

We evaluate the anomaly detection performance of the proposed model and the baseline models using three metrics: precision, recall, and F1 score. To detect anomalies, we obtain a threshold $\delta$ by labeling a proportion $r$ of the validation dataset as the anomaly set. Anomalies are identified if their anomaly scores at a given time point (Equation 11) exceed this threshold. Additionally, for the main results, we set $r$=0.5% for the SMD dataset, $r$=0.1% for the SWAT dataset, and $r$=1% for all other datasets.

Additionally, we follow the point adjustment strategy proposed in [40], which assumes that all anomalies within a continuous time period are correctly detected if any single point in that period is identified as anomalous. This strategy is reasonable because the detected anomalous point will trigger an alert, thereby drawing attention to the entire segment in practical applications.

## 6.4 Implementation Details

We conduct experiments on five real-world public datasets. To prevent overfitting, we separate 20% of samples from the training set as a validation set. Following the refined scheme proposed by Shen et al. [15], we use non-overlapping sliding windows to obtain a set of subsequences. For all datasets, the size of the sliding window is fixed at 100. The proposed MAD-transformer consists of a total of 3 layers, with the sequence and temporal state modules having non-shared parameters. We set their heads to 8 and the number of channels in the hidden state $d_{model}$ to 512. We empirically set $\lambda$ (Equation 11) to 19 for all datasets to balance the two parts of the loss function. We use ADAM [41] as the optimizer, with an initial learning rate of 1e-4, a batch size of 64, and the training process is early stopped after 10 epochs. For anomaly diagnosis, we follow MSCRED [9] to set the threshold $\tau = \beta \cdot \max\{s(t)_{\text{valid}}\}$, where $s(t)_{\text{valid}}$ are the anomaly scores over the validation period and $\beta \in [1,2]$ is set to maximize the F1 Score over the validation period. All experiments are implemented using a single NVIDIA A100-SXM4-80GB GPU in PyTorch.

## 6.5 Results and Analysis

Table 2: Comparison of anomaly detection results on different datasets

| Dataset | SMD | | | MSL | | | SMAP | | | SWaT | | | PSM | | | Avg |
|---|---|---|---|---|---|---|---|---|---|---|---|---|---|---|---|---|
| Metric | F1 | Pre | Rec | F1 | Pre | Rec | F1 | Pre | Rec | F1 | Pre | Rec | F1 | Pre | Rec | F1 (%) |
| OCSVM | 56.19 | 44.34 | 76.72 | 70.82 | 59.78 | 86.87 | 56.34 | 53.85 | 59.07 | 47.23 | 45.39 | 49.22 | 70.67 | 62.75 | 80.89 | 60.25 |
| IsolationForest | 53.64 | 42.31 | 73.29 | 66.45 | 53.94 | 86.54 | 55.53 | 52.39 | 59.07 | 47.02 | 49.29 | 44.95 | 83.48 | 76.09 | 92.45 | 61.22 |
| LOF | 46.68 | 56.34 | 39.86 | 61.18 | 47.72 | 85.25 | 57.60 | 58.93 | 56.33 | 68.62 | 72.15 | 65.43 | 70.61 | 57.89 | 90.49 | 60.94 |
| Deep-SVDD | 79.10 | 78.54 | 79.67 | 83.58 | 91.9 2 | 76.63 | 69.04 | 89.93 | 56.02 | 82.39 | 80.42 | 84.45 | 90.73 | 95.41 | 86.49 | 80.97 |
| DAGMM | 57.30 | 67.30 | 49.89 | 74.62 | 89.60 | 63.93 | 68.51 | 86.45 | 56.73 | 70.40 | 89.92 | 57.84 | 80.08 | 93.49 | 70.03 | 70.18 |
| MMPCACD | 75.02 | 71.20 | 79.28 | 69.95 | 81.42 | 61.31 | 81.73 | 88.61 | 75.84 | 74.73 | 82.52 | 68.29 | 77.29 | 76.26 | 78.35 | 75.74 |
| VAR | 74.08 | 78.35 | 70.26 | 77.90 | 74.68 | 81.42 | 64.83 | 81.38 | 53.88 | 69.34 | 81.59 | 60.29 | 87.13 | 90.71 | 83.82 | 74.66 |
| LSTM | 81.78 | 78.55 | 85.28 | 83.95 | 85.45 | 82.50 | 83.39 | 89.41 | 78.13 | 84.69 | 86.15 | 83.27 | 82.80 | 76.93 | 89.64 | 83.32 |
| CL-MPPCA | 79.09 | 82.36 | 76.07 | 80.44 | 73.71 | 88.54 | 72.88 | 86.13 | 63.16 | 79.07 | 76.78 | 81.50 | 71.80 | 56.02 | 99.93 | 76.66 |
| ITAD | 79.48 | 86.22 | 73.71 | 76.07 | 69.44 | 84.09 | 73.85 | 82.42 | 66.89 | 57.08 | 63.13 | 52.08 | 68.13 | 72.80 | 64.02 | 70.92 |
| LSTM-VAE | 82.30 | 75.76 | 90.08 | 82.62 | 85.49 | 79.94 | 78.10 | 92.20 | 67.75 | 82.20 | 76.00 | 89.50 | 80.96 | 73.62 | 89.92 | 81.24 |
| BeatGAN | 78.10 | 72.90 | 84.09 | 87.53 | 89.75 | 85.42 | 69.61 | 92.38 | 55.85 | 73.92 | 64.01 | 87.46 | 92.04 | 90.30 | 93.84 | 80.24 |
| OmniAnomaly | 85.22 | 83.68 | 86.82 | 87.67 | 89.02 | 86.37 | 86.92 | 92.49 | 81.99 | 82.83 | 81.42 | 84.30 | 80.83 | 88.39 | 74.46 | 84.69 |
| InterFusion | 86.22 | 87.02 | 85.43 | 86.62 | 81.28 | 92.70 | 89.14 | 89.77 | 88.52 | 83.01 | 80.59 | 85.58 | 83.52 | 83.61 | 83.45 | 85.70 |
| THOC | 84.99 | 79.76 | 90.95 | 89.69 | 88.45 | 90.97 | 90.68 | 92.06 | 89.34 | 85.13 | 83.94 | 86.36 | 89.54 | 88.14 | 90.99 | 88.01 |
| MAD-GAN | 85.10 | 85.96 | 84.25 | 91.38 | 85.55 | 98.07 | 88.14 | 94.22 | 82.79 | 86.53 | 90.85 | 82.60 | 87.90 | 88.25 | 87.56 | 87.81 |
| MSCRED | 78.75 | 84.73 | 79.11 | 82.05 | 87.37 | 77.34 | 69.33 | 94.11 | 54.88 | 84.59 | 92.69 | 77.80 | 83.78 | 87.23 | 80.60 | 79.7 |
| MTAD-GAT | 91.29 | 90.24 | 92.36 | 90.84 | 87.54 | 94.40 | 90.13 | 89.06 | 91.23 | 85.50 | 98.24 | 75.69 | 88.41 | 97.95 | 80.56 | 89.23 |
| GDN | **92.59** | 91.08 | 94.16 | **94.62** | 94.91 | 94.34 | 92.31 | 91.64 | 92.98 | 89.57 | 96.14 | 83.84 | 93.59 | 95.67 | 91.59 | 92.54 |
| USAD | 91.62 | 89.14 | 94.25 | 92.72 | 88.10 | 97.86 | 86.34 | 76.97 | 98.31 | 84.60 | 98.70 | 74.02 | 86.99 | 96.89 | 78.92 | 88.45 |
| Atransformer | **92.33** | 89.40 | 95.45 | 93.35 | 92.00 | 94.73 | **96.25** | 93.61 | 99.05 | **94.07** | 91.55 | 96.73 | **97.89** | 96.91 | 98.90 | 94.78 |
| Ours | 92.00 | 90.15 | 93.93 | **94.92** | 91.98 | 98.06 | **96.94** | 95.02 | 98.96 | **95.83** | 94.65 | 97.04 | **98.18** | 97.45 | 98.92 | **95.57** |

### 6.5.1 Anomaly Detection

The anomaly detection results of different models are shown in Table 2, where the best results are highlighted in red bold and the suboptimal in blue bold. Since F1 score balances precision and recall, we choose to analyze the model performance based on the F1 score. A higher F1 score indicates better performance.

1) On five public datasets: We conducted extensive experiments on five real-world public datasets with 21 benchmark models. As shown in Table 2, overall, we found that the model proposed in this paper outperforms all the benchmark models.

We observed that (A) MTAD-GAT, GDN, and Atransformer achieved relatively better performance than most baseline methods, which validates the effectiveness of modeling spatial dependencies in time series and the associations within the adjacency set for anomaly detection. (B) The proposed method still outperformed all three in almost all setups, demonstrating the clear superiority of explicitly modeling sequence associations, temporal dependencies, and spatial dependencies in enhancing the performance of anomaly detection.

2) CTCS dataset: This dataset was collected on a petroleum refining simulation platform that we built, which couples a tank control system. The simulation platform, as illustrated in the figure, includes seven sensors and two water tanks. The detection results on this dataset are shown in Figure 5, and the proposed model still achieves state-of-the-art performance. This further validates the effectiveness of the model.

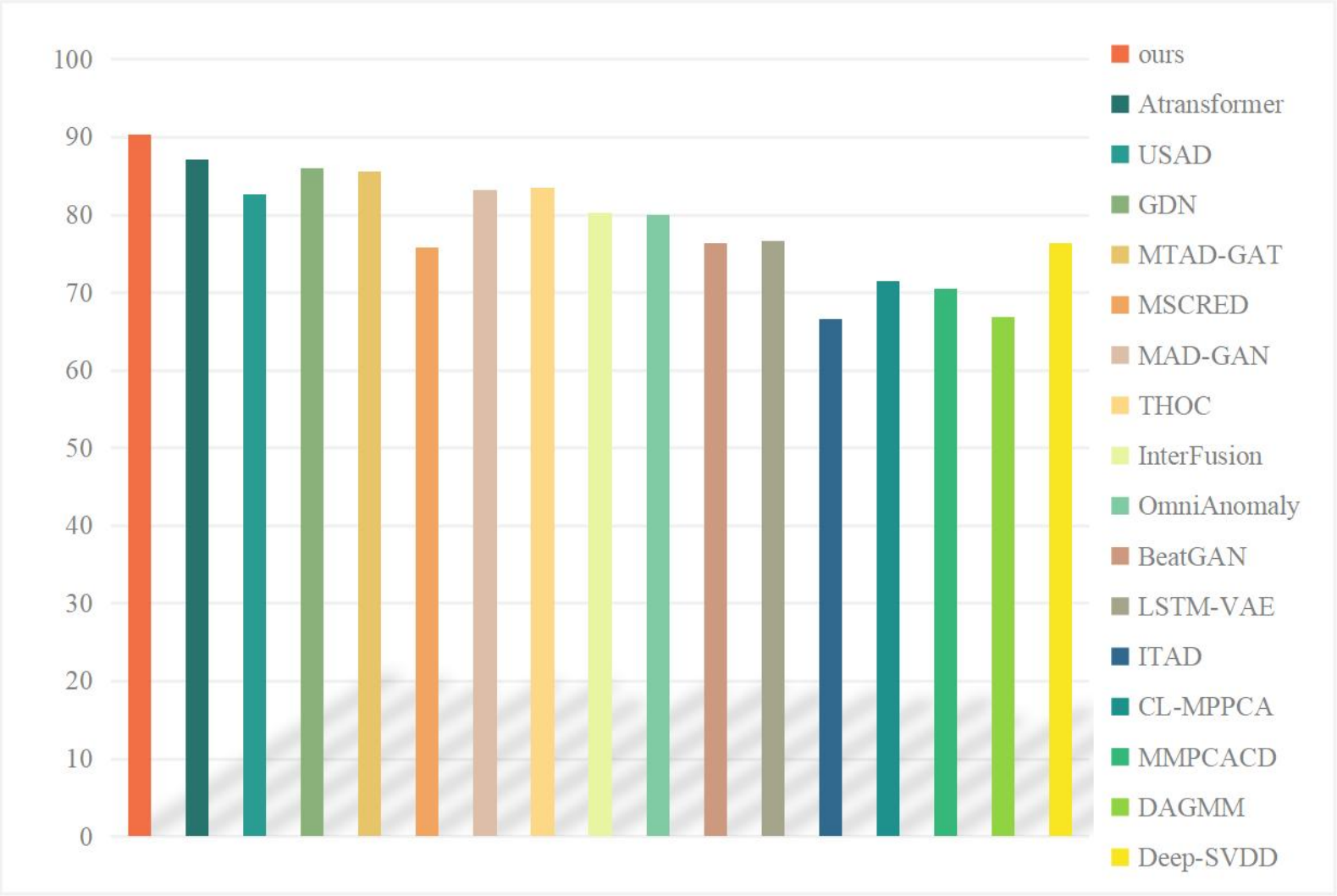

Figure 5. Results for CTCS dataset.

3) To present a more detailed comparison, we compared the proposed model and two best-performing baseline methods (i.e., Atransformer and GDN) on the CTCS dataset. We can observe that the anomaly scores of GDN are unstable, with many false positives and false negatives in the results. At the same time, the anomaly scores of Atransformer are smoother than GDN's, but there are still some false positives and false negatives. The MAD-Transformer, however, detects all anomalies without any false positives or false negatives. This further validates the effectiveness of the proposed model in detecting anomalies by comprehensively modeling sequence associations, temporal dependencies, and spatial dependencies, resulting in a significant improvement in detection performance.

### 6.5.2 Ablation Study

As shown in Table 3, we further investigated the impact of each component within the model. The table reveals that each branch of the proposed MAD-Attention can further enhance the model's performance. Specifically, modeling temporal dependencies significantly improved the average absolute F1 score by 5.01% (90.56 → 95.57), modeling spatial dependencies significantly improved by 3.88% (91.69 → 95.57), and modeling sequence dependencies significantly improved by 5.91% (89.66 → 95.57). Additionally, the designed alignment-based loss also enhanced the model's performance (89.66 → 95.57). Ultimately, the proposed MAD-Transformer shows a 17.02% absolute improvement over the pure Transformer (78.55 → 95.57). These results validate that each module in our design is necessary and effective.

Table 3. Ablation study results (F1-score). “w.o” means without and “w.o align loss” means without the association alignment loss.

| Architecture | SMD | MSL | SMAP | SWAT | PSM | Avg F1(%) |
|---|---|---|---|---|---|---|
| MAD-Transformer | 92.00 | 94.92 | 96.94 | 95.83 | 98.18 | 95.57 |
| w.o temporal branch | 88.25 | 89.06 | 90.89 | 90.25 | 94.34 | 90.56 |
| w.o spatial branch | 89.65 | 90.98 | 92.31 | 91.56 | 93.97 | 91.69 |
| w.o series branch | 86.39 | 88.56 | 90.68 | 90.05 | 92.61 | 89.66 |
| w.o align loss | 90.96 | 92.85 | 94.64 | 92.89 | 95.59 | 93.39 |
| transformer | 76.89 | 80.66 | 79.53 | 78.72 | 76.94 | 78.55 |

### 6.5.3 Anomaly Localization

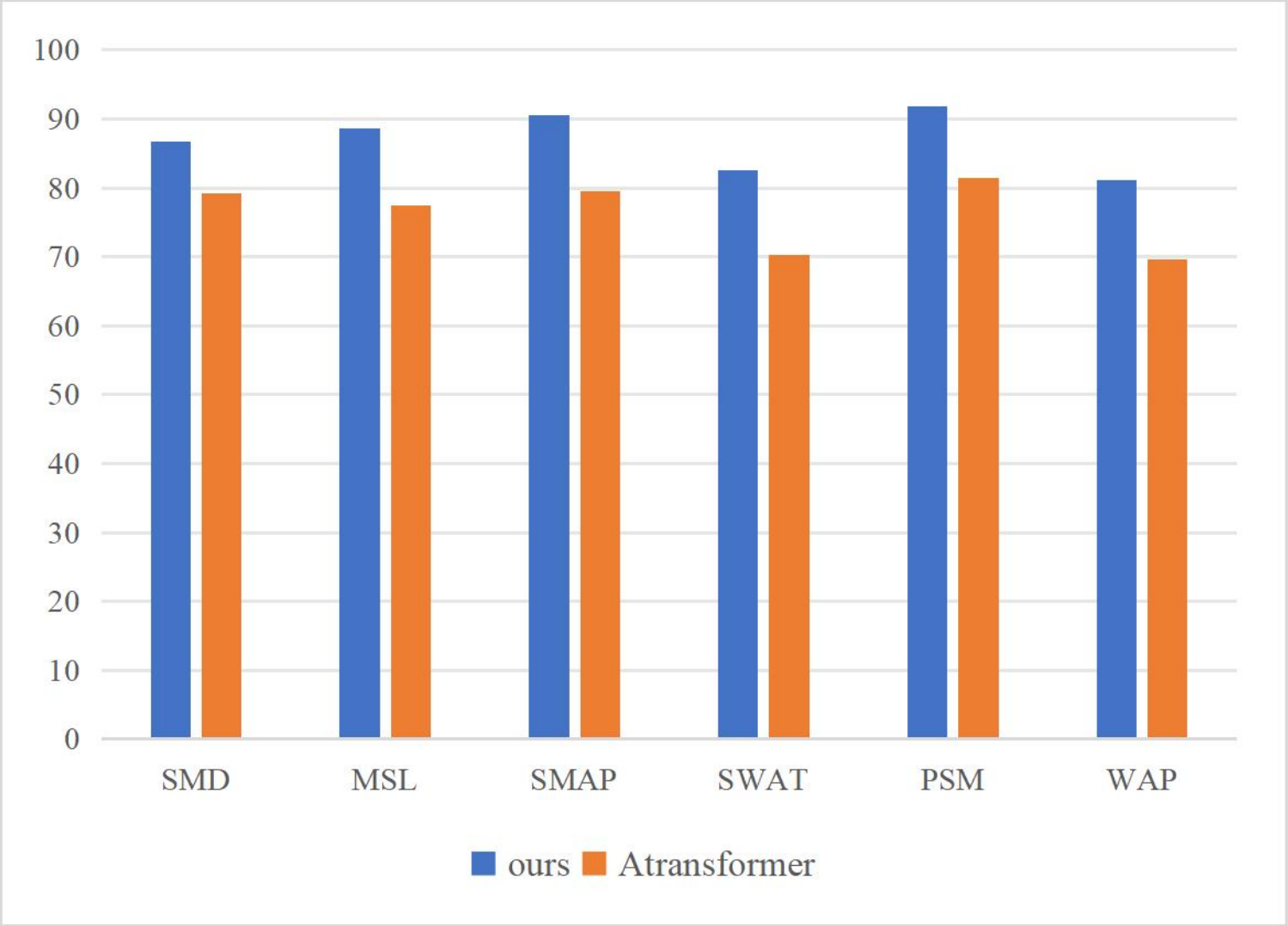

Figure 6. Performance of anomaly localization

Abnormal localization plays a crucial role in CPS systems, capable of quickly identifying abnormal behaviors or faults within the system, thereby taking measures to prevent and reduce potential security threats. This is of profound significance for ensuring the safety, reliability, and efficient operation of the system. As one of the tasks in abnormal diagnosis, abnormal localization relies on good abnormal detection performance. Therefore, we compared the performance of MAD-Transformer with the best baseline, Atransformer. Specifically, unlike Atransformer, which uses the reconstruction error of each time series as the abnormal score for that sequence, the abnormal score of MAD-Transformer is defined as the number of poor reconstructed pairwise correlations in a specific row or column of the residual state matrix. This is because each row/column represents a time series (device). For each abnormal event, we sort all time series based on their abnormal scores and identify the top k sequences as the root causes. Figure 6 shows the average recall@k (k=3) from 5 repeated experiments. On the six datasets, the performance of MAD-Transformer is higher than Atransformer by 12.8% on average.

### 6.5.4 Fine-grained Anomaly Diagnosis

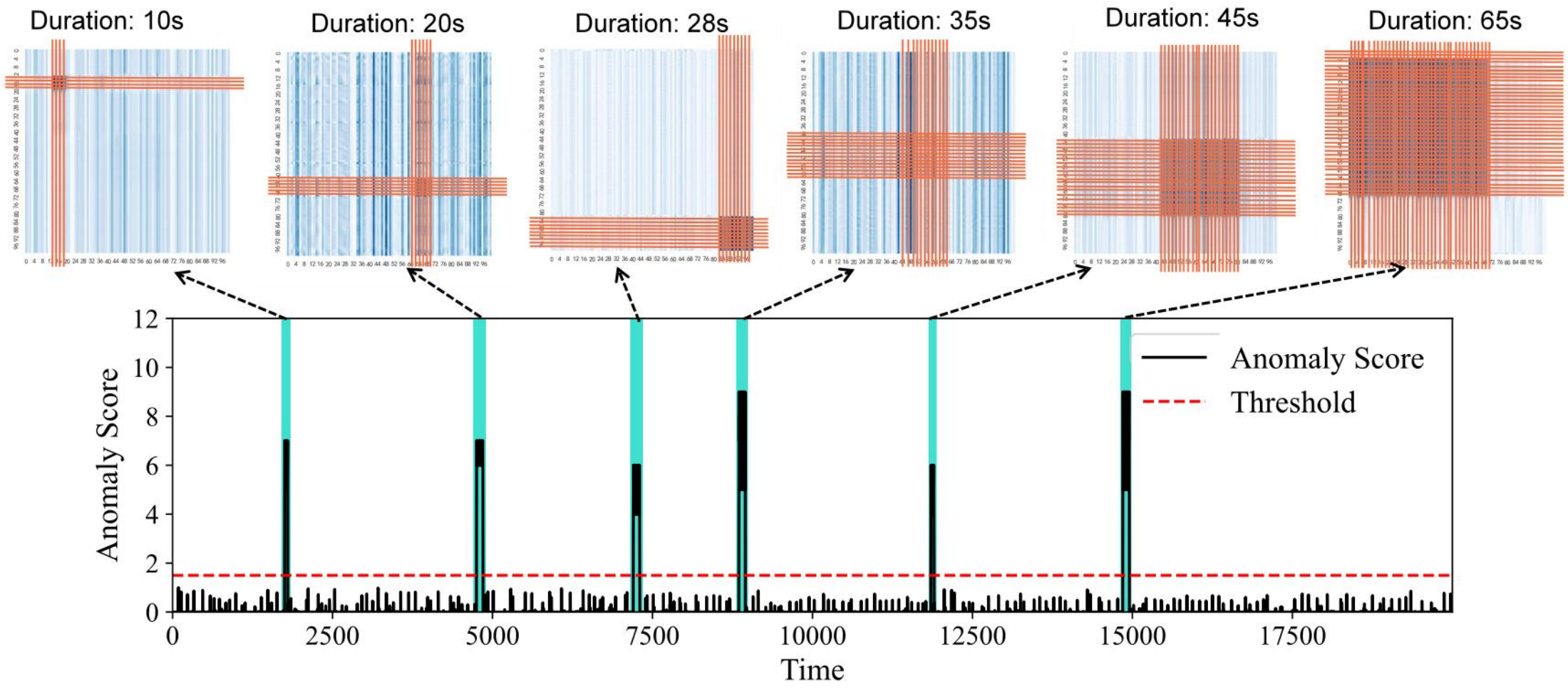

Figure 7. Case study of anomaly diagnosis

The number of rows in the model's time feature matrix corresponds to the size of the sliding window (win_size=100), which can capture the system state for the duration of win_size. To explain the severity of anomalies, we first calculate the anomaly score for each row based on the time residual feature matrix. Each row represents a specific time point, so the model can assess the anomalous performance persisting for any duration within the configurable win_size.

To evaluate and demonstrate the effectiveness of the model in detecting and diagnosing anomalies, Figure 7 presents a case study of abnormal diagnosis on a self-collected real dataset CTCS. Notably, the dataset contains six types of anomalies, each lasting for a duration of 10, 20, 30, 40, 50, and 60 seconds, respectively. In this case, the proposed model is capable of detecting all six types of anomalies. Furthermore, the six time residual feature matrices injected with abnormal events provide diagnostic results for the duration of anomalies. Anomalies in the matrix are highlighted in red, with each red row/column representing an anomalous time step. The greater the number of red rows/columns, the longer the duration of the anomaly. As illustrated in Figure 7, the proposed model can identify the duration of each anomaly with an accuracy of approximately 92%. The identified anomalies are sorted by duration (i.e., severity) as A1 > A2 > A3 > A4, consistent with the predefined results. Therefore, in this case, we can accurately assess the severity of each anomaly.

Although the model MSCRED can also determine the severity of anomalies, it relies on three different scales of spatial state matrices calculated during training, with each residual matrix computing an anomaly score. Consequently, it can only identify anomalies that persist for a duration equal to or greater than the scale of the matrix and cannot compute anomalies at each time point. Therefore, it can only roughly assess the severity of the predefined three scales of anomalies; if the duration is less than the corresponding scale, it fails to recognize the anomaly. For example, anomaly 5 persists for 50 seconds. Our model can roughly identify the duration of the anomaly as 45 time steps, as shown in matrix 5. However, due to the limitation of its three scales of spatial state matrices, MSCRED can only determine anomalies that persist for more than 30 time steps and cannot accurately assess how many time steps the anomaly actually lasts. This further demonstrates the superiority of our model in diagnosing the severity of anomalies.

## 7. Conclusion

For the task of unsupervised time series anomaly detection, based on the analysis of correlations within time series, temporal state matrices and spatial state matrices are built to model the state correlations of industrial CPS respectively across different time periods. We propose a MAD-Transformer with a three-branch structure that explicitly captures the series-temporal-spatial dependencies among MTS sensors. Furthermore, temporal residual feature matrices and spatial residual feature matrices are utilized to adaptively locate and diagnose anomalies. Extensive empirical studies on five widely used public datasets and a self-collected petroleum refining simulation dataset have shown that the proposed MAD-transformer not only outperforms the most advanced baseline methods in terms of noise robustness and localization performance, but can also adaptively detect fine-grained anomalies with short duration.

## Acknowledgement

This work was supported by the National Key Research and Development Program of China (No. 2022YFB3103402) and the National Natural Science Foundation of China (No. 62072200, No. 62172176, No. 62127808).